%% file: main.tex
\documentclass[10pt,twocolumn,letterpaper]{article}

\usepackage{cvpr}              
\usepackage{multirow}
\usepackage{bbm}
\usepackage{makecell}
\usepackage{tabularx}

\input{preamble}

\definecolor{cvprblue}{rgb}{0.21,0.49,0.74}
\usepackage[pagebackref,breaklinks,colorlinks,citecolor=cvprblue]{hyperref}

\definecolor{cvprblue}{rgb}{0.21,0.49,0.74}

\def\paperID{10972} 
\def\confName{CVPR}
\def\confYear{2024}

\def\Approach{Stable Messenger}

\definecolor{mydarkblue}{rgb}{0,0.08,1}
\definecolor{mydarkgreen}{rgb}{0.02,0.6,0.02}
\definecolor{myred}{rgb}{1.0,0.0,0.0}
\definecolor{myred2}{rgb}{0.7,0.1,0.1}
\definecolor{mydarkblue2}{rgb}{0.05,0.1,0.7}
\definecolor{mypurple}{rgb}{111,0,255}
\definecolor{gray}{rgb}{0.3, 0.3, 0.3}

\title{\Approach: Steganography for Message-Concealed Image Generation}

\author{%
 Quang Ho Nguyen$^{1}$\thanks{First two authors contribute equally. The work is done during Quang Nguyen's internship at VinAI Research.} \quad Truong Vu$^{1}$\footnotemark[1] \quad Cuong Pham$^{1,2}$ \quad Anh Tran$^{1}$ \quad Khoi Nguyen$^{1}$ \\ \\
  $^1$VinAI Research, \\ $^2$Posts \& Telecommunications Inst. of Tech., Vietnam
 \\
}

\begin{document}
\input{definitions}
\maketitle
\input{sec/0_abstract}

\input{sec/1_intro}
\input{sec/2_related}

\input{sec/3_approach}

\input{sec/4_experiments}

\input{sec/5_conclusion}

{
    \small
    \bibliographystyle{ieeenat_fullname}
    \bibliography{main}
}



\end{document}


\input{definitions}
\maketitle

In this supplementary material, we first elaborate on the implementation details in Sec.~\ref{sec:implementation}.
Next, we show more ablation studies in Sec.~\ref{sec:more_ablation}.
Then, we present results with the longer messages in Sec.~\ref{sec:quantitative}.
Finally, we provide more qualitative results in Sec.~\ref{sec:more-qualitative}. 

\section{Implementation Details}
\label{sec:implementation}


We visualize the architecture of our \textbf{latent-aware message encoder} $E_m$ in Fig.~\ref{fig:message-encoder}. Our message decoder $D_m$ is ResNet50, with the output layer modified to fit the number of bits considered. All experiments are conducted with a single A100 40GB GPU. 
The training duration for our network varies depending on the length of the message. Specifically, the training for the 100-bit setting spans approximately three days until convergence, while that of the 150-bit and 200-bit settings is more than five days. 

\begin{figure}[h]
    \centering
    \includegraphics[width=1\linewidth]{figures/message_encoder_diagram}
    \caption{Architecture of latent-aware message encoder}
    \label{fig:message-encoder}
\end{figure}

\noindent Also, we set the \textbf{parameters for the transformations} in the robust evaluation (Sec. 4.3 in the main paper) as follows:
\begin{itemize}
    \item Gaussian blur: kernel size is 5, and $\sigma = 2$
    \item Gaussian noise: $(\mu, \sigma) = $ (0, 0.2)
    \item JPEG Compression: keep 80\% image quality.
\end{itemize}




\section{More Ablation Studies}
\label{sec:more_ablation}

We study the effect of different versions of Stable Diffusion (SD) in Tab.~\ref{tab:sd_versions} on MirFlickr with ours+LSE. As can be seen, all versions yield very similar results. 
Also, we study more about the bit accuracy threshold $\tau_2$ used to activate the LSE loss in Tab.~\ref{tab:activate_lse}. We observe that the threshold of $\tau_2=95\%$ gives the best results. 

\begin{table}[t]
    \small
    \centering
    \begin{tabular}{lccc}
    \toprule
         \bf SD Version     &   \bf PSNR    & \bf Bit Acc & \bf Message Acc   \\
         \toprule
         SD v1.4        &   35.15       &99\%          &88\%            \\  
         SD v1.5        &   35.18       &99\%          &89\%              \\ 
         SD v2.1        &   35.36       &99\%          &88\%              \\
    \toprule
    \end{tabular}
    \vspace{-10pt}
    \caption{Results of SD versions in generative mode (with 100 bits).}
    \vspace{-5pt}
    \label{tab:sd_versions}
\end{table}

\begin{table}[t]
    \small
    \centering
    \begin{tabular}{cccc}
    \toprule
        \textbf{Threshold $\tau_2$} &  \bf PSNR    &   \bf Bit Acc     & \bf Message Acc  \\
        \toprule
         50\% &                 33.07           & 99\%           & 68\%          \\
         80\%   &33.49                          &99\%              &90\%          \\
         95\%   &33.11                       &99\%              &94\%          \\
         \toprule
    \end{tabular}
    \vspace{-5pt}
    \caption{Study on the threshold to activate LSE (with 100 bits).}
    \vspace{-10pt}
    \label{tab:activate_lse}
\end{table}

\section{Results with Longer Messages}
\label{sec:quantitative}

We present the results with the longer messages in Tab.~\ref{tab:comparison_150} (150 bits) and Tab.~\ref{tab:comparison_200} (200 bits). Firstly, the LSE loss consistently enhances message accuracy, particularly when the original accuracy exceeds 0, validating the efficacy of this proposed loss function. Occasionally, with 200-bit messages, both StegaStamp \cite{tancik2020stegastamp} and RoSteALS \cite{bui2023rosteals} fail, resulting in 0\% message accuracy despite the application of our LSE loss. Secondly, in comparisons among steganographic architectures, RoSteALS \cite{bui2023rosteals} consistently yields the lowest message accuracy. StegaStamp \cite{tancik2020stegastamp} outperforms our network in message accuracy with a 150-bit setup across all test datasets. However, when assessed with 200 bits, our approach maintains acceptable message accuracy while StegaStamp's accuracy drops significantly, nearing zero. As a result, our \Approach~consistently preserves message accuracy more effectively over longer messages. Additionally, StegaStamp \cite{tancik2020stegastamp} exhibits the lowest PSNR scores, whereas our method offers a favorable balance between message accuracy and stealthiness.



\begin{table*}[t]
\small
    \centering
    \begin{tabular}{llccccl}
    \toprule
     \bf Datasets & \bf Methods & \bf PSNR \(\uparrow \)& \bf SSIM \(\uparrow\)& \bf LPIPS \( \downarrow\) & \bf Bit Acc (\%) \(\uparrow\) & \bf Message Acc (\%) \(\uparrow\) \\ 
    \toprule 
    \multirow{ 6}{*}{\textbf{MirFlickr \cite{huiskes2008mir}}} & StegaStamp \cite{tancik2020stegastamp}   & 32.36     &0.92      &  0.07                      & 99           &77    \\
    & StegaStamp \cite{tancik2020stegastamp} + LSE   & 32.33  & 0.92    &  0.07                      &  99       &      94 \textbf{(+17)}            \\
     & RoSteALS \cite{bui2023rosteals}      & 33.14     & 0.93      &0.04 & 98  & 20        \\
    & RoSteALS \cite{bui2023rosteals} + LSE     &  33.10     &0.93   &  0.05   &   99     & 42 \textbf{(+22)}                \\
    & Ours  & 33.30 & 0.87  & 0.08 & 99 & 39\\
    & Ours + LSE    &33.15 &0.87  &0.08  &99  & 51 \textbf{(+12)}  \\
    \toprule 
    \multirow{ 6}{*}{\textbf{CLIC \cite{mari2022clic}}} & StegaStamp \cite{tancik2020stegastamp}    &32.47    &0.95     &0.07                         &  99          &      82\\
    & StegaStamp \cite{tancik2020stegastamp} + LSE   & 32.47  &0.95    & 0.07                      & 99        &       94 \textbf{(+12)}   \\
     & RoSteALS \cite{bui2023rosteals}      &33.35    & 0.97    & 0.04 & 98  & 20             \\
    & RoSteALS \cite{bui2023rosteals} + LSE     & 33.31    & 0.97    & 0.04 & 99   & 43 \textbf{(+23)}                  \\
    & Ours          &33.55  &0.93   & 0.09  & 99    &41\\ 
    & Ours  + LSE   &33.48  &0.93  &0.09  & 99  & 56 \textbf{(+15)}  \\
    \toprule 
    \multirow{ 6}{*}{\textbf{MetFaces \cite{karras2020metfaces}}} & StegaStamp \cite{tancik2020stegastamp}    & 33.57     &0.95          & 0.12                          &   99            &     92             \\
    & StegaStamp \cite{tancik2020stegastamp} + LSE   & 33.66    &0.95    & 0.13                        &  99       &       97 \textbf{(+5)}           \\
     & RoSteALS \cite{bui2023rosteals}      &35.80  &0.97     &0.06     &98     &17               \\
    & RoSteALS \cite{bui2023rosteals} + LSE     & 35.73      &0.97      &0.07   & 99   & 43 \textbf{(+26) }                 \\
    & Ours        &34.95  &0.91     &0.13   &99 &50\\
    & Ours + LSE  &34.70  &0.91  &0.13  & 99  & 59 \textbf{(+9)} \\ 
    \bottomrule
    \end{tabular} 
    \caption{Results of our \Approach~and prior work with and without the LSE loss on various datasets using \textbf{150-bit message}.
    }
    \label{tab:comparison_150}
\end{table*}

\begin{table*}[t]
\small
    \centering
    \begin{tabular}{llccccl}
    \toprule
     \bf Datasets & \bf Methods & \bf PSNR \(\uparrow \)& \bf SSIM \(\uparrow\)& \bf LPIPS \( \downarrow\) & \bf Bit Acc (\%) \(\uparrow\) & \bf Message Acc (\%) \(\uparrow\) \\ 
    \toprule 
    \multirow{ 6}{*}{\textbf{MirFlickr \cite{huiskes2008mir}}} & StegaStamp \cite{tancik2020stegastamp}   &32.45    &0.92     &0.07                         &  95          &      0 \\
    & StegaStamp \cite{tancik2020stegastamp} + LSE  & 32.44  &0.92    & 0.07                      & 96        &       1 \textbf{(+1)}          \\
     & RoSteALS \cite{bui2023rosteals}      & 33.42     & 0.93      &0.04 & 94  & 0              \\
    & RoSteALS \cite{bui2023rosteals} + LSE     &  33.36     &0.93   &  0.05   &   95     & 0                 \\
    & Ours        &33.48    &0.87   &0.06  &99  &30 \\
    & Ours + LSE  &33.51  &0.87  &0.06  & 99  & 39 \textbf{(+9)}\\ 
    
    \toprule 
    \multirow{ 6}{*}{\textbf{CLIC \cite{mari2022clic}}} & StegaStamp \cite{tancik2020stegastamp}    & 32.57    &0.95     &0.06                         &  96          &      0\\
    & StegaStamp \cite{tancik2020stegastamp} + LSE    & 32.53    &0.94    & 0.07  &  98       & 0  \\
     & RoSteALS \cite{bui2023rosteals}      &33.69    & 0.97    & 0.04 & 94  & 0         \\
    & RoSteALS \cite{bui2023rosteals} + LSE     & 33.63    & 0.97    & 0.04 & 96   & 0                \\
    & Ours          & 34.04 & 0.94  &0.08   &99 &32\\
    & Ours  + LSE   & 33.89 &0.94  &0.08  &99  &43 \textbf{(+11)}  \\
    \toprule 
    \multirow{ 6}{*}{\textbf{MetFaces \cite{karras2020metfaces}}} & StegaStamp \cite{tancik2020stegastamp}    & 33.64     & 0.96     &  0.12                      & 96           &0            \\
    & StegaStamp \cite{tancik2020stegastamp} + LSE   & 32.53    &0.94    & 0.07                        &  98       &       0            \\
     & RoSteALS \cite{bui2023rosteals}      &36.30  &0.98     &0.06 &94 &0               \\
& RoSteALS \cite{bui2023rosteals} + LSE     & 36.16      &0.97      &0.07   & 95   & 0        \\
    & Ours        &35.82    &0.92   &0.10   &99 & 41\\
    & Ours + LSE  &35.57  &0.92  &0.10  &99   &51  \textbf{(+10)}\\ 
    \bottomrule
    \end{tabular} 
    \caption{Results of our \Approach~and prior work with and without the LSE loss on various datasets using \textbf{200-bit message}.
    }
    \vspace{-10pt}
    \label{tab:comparison_200}
\end{table*}

\section{More Qualitative Results}
\label{sec:more-qualitative}

We present qualitative results for the cover mode in Figs. \ref{fig:qualitative-mir}-\ref{fig:qualitative_metfaces} and the generative mode in Fig. \ref{fig:supp-generative-mode}. Our method sustains high image quality in both modes, occasionally showing artifacts in white regions.

\begin{figure*}[t]
    \centering
    \includegraphics[width=1\linewidth]{figures/qualitative_mir5.pdf}
    \caption{\textbf{Qualitative results in the cover mode on the MirFlickr dataset}. Our results are displayed in two columns, each row presenting an example: the cover image, steganographic image, and their difference from left to right. While \Approach~effectively conceals the hidden message, occasional artifacts, particularly within the \textcolor{red}{red ovals} in the right column, may be noticeable.}
    \label{fig:qualitative-mir}
\end{figure*}

\begin{figure*}[t]
    \centering
    \includegraphics[width=1\linewidth]{figures/qualitative_clic5.pdf}
    \caption{\textbf{Qualitative results in the cover mode on the CLIC dataset}. See the caption of Fig.~\ref{fig:qualitative-mir} for more details.}
    \label{fig:qualitative-clic}
\end{figure*}

\begin{figure*}[t]
    \centering
    \includegraphics[width=1\linewidth]{figures/qualitative_metfaces3.pdf}
    \caption{\textbf{Qualitative results in the cover mode on the Metfaces dataset}. See the caption of Fig.~\ref{fig:qualitative-mir} for more details.}
    \label{fig:qualitative_metfaces}
\end{figure*}

\begin{figure*}[t]
    \centering
    \includegraphics[width=1.0\linewidth]{figures/generative_mode6.pdf}
    \caption{\textbf{Qualitative results in the generative mode}. Each example includes a generated image paired with its input text prompt. \Approach~successfully generates high-fidelity steganographic images (first three rows) but occasionally reveals artifacts (last row).}
    \label{fig:supp-generative-mode}
\end{figure*}

{
    \small
    \bibliographystyle{ieeenat_fullname}
    \bibliography{main}
}


%% file: preamble.tex
\usepackage[dvipsnames]{xcolor}


%% file: definitions.tex
\def\mA{\mathcal{A}}
\def\mB{\mathcal{B}}
\def\mC{\mathcal{C}}
\def\mD{\mathcal{D}}
\def\mE{\mathcal{E}}
\def\mF{\mathcal{F}}
\def\mG{\mathcal{G}}
\def\mH{\mathcal{H}}
\def\mI{\mathcal{I}}
\def\mJ{\mathcal{J}}
\def\mK{\mathcal{K}}
\def\mL{\mathcal{L}}
\def\mM{\mathcal{M}}
\def\mN{\mathcal{N}}
\def\mO{\mathcal{O}}
\def\mP{\mathcal{P}}
\def\mQ{\mathcal{Q}}
\def\mR{\mathcal{R}}
\def\mS{\mathcal{S}}
\def\mT{\mathcal{T}}
\def\mU{\mathcal{U}}
\def\mV{\mathcal{V}}
\def\mW{\mathcal{W}}
\def\mX{\mathcal{X}}
\def\mY{\mathcal{Y}}
\def\mZ{\mathcal{Z}} 

\def\bbN{\mathbb{N}} 
\def\bbR{\mathbb{R}} 
\def\bbP{\mathbb{P}} 
\def\bbQ{\mathbb{Q}} 
\def\bbE{\mathbb{E}}

\def\1n{\mathbf{1}_n}
\def\0{\mathbf{0}}
\def\1{\mathbf{1}}

\def\A{{\bf A}}
\def\B{{\bf B}}
\def\C{{\bf C}}
\def\D{{\bf D}}
\def\E{{\bf E}}
\def\F{{\bf F}}
\def\G{{\bf G}}
\def\H{{\bf H}}
\def\I{{\bf I}}
\def\J{{\bf J}}
\def\K{{\bf K}}
\def\L{{\bf L}}
\def\M{{\bf M}}
\def\N{{\bf N}}
\def\O{{\bf O}}
\def\P{{\bf P}}
\def\Q{{\bf Q}}
\def\R{{\bf R}}
\def\S{{\bf S}}
\def\T{{\bf T}}
\def\U{{\bf U}}
\def\V{{\bf V}}
\def\W{{\bf W}}
\def\X{{\bf X}}
\def\Y{{\bf Y}}
\def\Z{{\bf Z}}

\def\a{{\bf a}}
\def\b{{\bf b}}
\def\c{{\bf c}}
\def\d{{\bf d}}
\def\e{{\bf e}}
\def\f{{\bf f}}
\def\g{{\bf g}}
\def\h{{\bf h}}
\def\i{{\bf i}}
\def\j{{\bf j}}
\def\k{{\bf k}}
\def\l{{\bf l}}
\def\m{{\bf m}}
\def\n{{\bf n}}
\def\o{{\bf o}}
\def\p{{\bf p}}
\def\q{{\bf q}}
\def\r{{\bf r}}
\def\s{{\bf s}}
\def\t{{\bf t}}
\def\u{{\bf u}}
\def\v{{\bf v}}
\def\w{{\bf w}}
\def\x{{\bf x}}
\def\y{{\bf y}}
\def\z{{\bf z}}

\def\balpha{\mbox{\boldmath{$\alpha$}}}
\def\bbeta{\mbox{\boldmath{$\beta$}}}
\def\bdelta{\mbox{\boldmath{$\delta$}}}
\def\bgamma{\mbox{\boldmath{$\gamma$}}}
\def\blambda{\mbox{\boldmath{$\lambda$}}}
\def\bsigma{\mbox{\boldmath{$\sigma$}}}
\def\btheta{\mbox{\boldmath{$\theta$}}}
\def\bomega{\mbox{\boldmath{$\omega$}}}
\def\bxi{\mbox{\boldmath{$\xi$}}}
\def\bnu{\mbox{\boldmath{$\nu$}}}                                  
\def\bphi{\mbox{\boldmath{$\phi$}}}
\def\bmu{\mbox{\boldmath{$\mu$}}}

\def\bDelta{\mbox{\boldmath{$\Delta$}}}
\def\bOmega{\mbox{\boldmath{$\Omega$}}}
\def\bPhi{\mbox{\boldmath{$\Phi$}}}
\def\bLambda{\mbox{\boldmath{$\Lambda$}}}
\def\bSigma{\mbox{\boldmath{$\Sigma$}}}
\def\bGamma{\mbox{\boldmath{$\Gamma$}}}
                                  
\newcommand{\myprob}[1]{\mathop{\mathbb{P}}_{#1}}

\newcommand{\myexp}[1]{\mathop{\mathbb{E}}_{#1}}

\newcommand{\mydelta}[1]{1_{#1}}

\newcommand{\myminimum}[1]{\mathop{\textrm{minimum}}_{#1}}
\newcommand{\mymaximum}[1]{\mathop{\textrm{maximum}}_{#1}}    
\newcommand{\mymin}[1]{\mathop{\textrm{minimize}}_{#1}}
\newcommand{\mymax}[1]{\mathop{\textrm{maximize}}_{#1}}
\newcommand{\mymins}[1]{\mathop{\textrm{min.}}_{#1}}
\newcommand{\mymaxs}[1]{\mathop{\textrm{max.}}_{#1}}  
\newcommand{\myargmin}[1]{\mathop{\textrm{argmin}}_{#1}} 
\newcommand{\myargmax}[1]{\mathop{\textrm{argmax}}_{#1}} 
\newcommand{\myst}{\textrm{s.t. }}

\newcommand{\denselist}{\itemsep -1pt}
\newcommand{\sparselist}{\itemsep 1pt}

\definecolor{pink}{rgb}{0.9,0.5,0.5}
\definecolor{purple}{rgb}{0.5, 0.4, 0.8}   
\definecolor{gray}{rgb}{0.3, 0.3, 0.3}
\definecolor{mygreen}{rgb}{0.2, 0.6, 0.2}

\newcommand{\cyan}[1]{\textcolor{cyan}{#1}}
\newcommand{\blue}[1]{\textcolor{blue}{#1}}
\newcommand{\magenta}[1]{\textcolor{magenta}{#1}}
\newcommand{\pink}[1]{\textcolor{pink}{#1}}
\newcommand{\green}[1]{\textcolor{green}{#1}} 
\newcommand{\gray}[1]{\textcolor{gray}{#1}}    
\newcommand{\mygreen}[1]{\textcolor{mygreen}{#1}}    
\newcommand{\purple}[1]{\textcolor{purple}{#1}}       

\definecolor{greena}{rgb}{0.4, 0.5, 0.1}
\newcommand{\greena}[1]{\textcolor{greena}{#1}}

\definecolor{bluea}{rgb}{0, 0.4, 0.6}
\newcommand{\bluea}[1]{\textcolor{bluea}{#1}}
\definecolor{reda}{rgb}{0.6, 0.2, 0.1}
\newcommand{\reda}[1]{\textcolor{reda}{#1}}

\def\changemargin#1#2{\list{}{\rightmargin#2\leftmargin#1}\item[]}
\let\endchangemargin=\endlist
                                               
\newcommand{\cm}[1]{}

\newcommand{\mhoai}[1]{{\color{magenta}\textbf{[MH: #1]}}}

\newcommand{\mtodo}[1]{{\color{red}$\blacksquare$\textbf{[TODO: #1]}}}
\newcommand{\myheading}[1]{\vspace{1ex}\noindent \textbf{#1}}
\newcommand{\htimesw}[2]{\mbox{$#1$$\times$$#2$}}


\newif\ifshowsolution
\showsolutiontrue

\ifshowsolution  
\newcommand{\Comment}[1]{\paragraph{\bf $\bigstar $ COMMENT:} {\sf #1} \bigskip}
\newcommand{\Solution}[2]{\paragraph{\bf $\bigstar $ SOLUTION:} {\sf #2} }
\newcommand{\Mistake}[2]{\paragraph{\bf $\blacksquare$ COMMON MISTAKE #1:} {\sf #2} \bigskip}
\else
\newcommand{\Solution}[2]{\vspace{#1}}
\fi

\newcommand{\truefalse}{
\begin{enumerate}
	\item True
	\item False
\end{enumerate}
}

\newcommand{\yesno}{
\begin{enumerate}
	\item Yes
	\item No
\end{enumerate}
}

\newcommand{\Sref}[1]{Sec.~\ref{#1}}
\newcommand{\Eref}[1]{Eq.~(\ref{#1})}
\newcommand{\Fref}[1]{Fig.~\ref{#1}}
\newcommand{\Tref}[1]{Table~\ref{#1}}

%% file: sec/0_abstract.tex
\begin{abstract}
In the ever-expanding digital landscape, safeguarding sensitive information remains paramount. This paper delves deep into digital protection, specifically focusing on steganography. While prior research predominantly fixated on individual bit decoding, we address this limitation by introducing ``message accuracy'', a novel metric evaluating the entirety of decoded messages for a more holistic evaluation. In addition, we propose an adaptive universal loss tailored to enhance message accuracy, named Log-Sum-Exponential (LSE) loss, thereby significantly improving the message accuracy of recent approaches. Furthermore, we also introduce a new latent-aware encoding technique in our framework named \Approach, harnessing pretrained Stable Diffusion for advanced steganographic image generation, giving rise to a better trade-off between image quality and message recovery. Throughout experimental results, we have demonstrated the superior performance of the new LSE loss and latent-aware encoding technique. This comprehensive approach marks a significant step in evolving evaluation metrics, refining loss functions, and innovating image concealment techniques, aiming for more robust and dependable information protection.
\end{abstract}

%% file: sec/1_intro.tex
\section{Introduction}

\label{sec:introduction}

Data hiding techniques, like watermarking and steganography, are crucial for securing sensitive information in our increasingly digital landscape. While both conceal messages within data, watermarking protects ownership, while steganography ensures secure communication. Images, with their rich presentation and widespread digital presence, serve as ideal carriers for hidden messages. Extensive research on digital image watermarking and steganography dates back to early computer vision \cite{zhang2004steganography,nguyen2006multi,mielikainen2006lsb,luo2010edge,alam2014key}. Recent advancements in deep learning-based methods \cite{zhu2018hidden, zhang1909rivagan} achieve highly accurate message injection and recovery while imperceptibly altering the host images. Notably, StegaStamp \cite{tancik2020stegastamp} introduces a robust steganographic approach capable of preserving hidden messages even through extreme image transformations, applicable in the real world.

\begin{figure}
    \centering
    \includegraphics[width=1\linewidth]{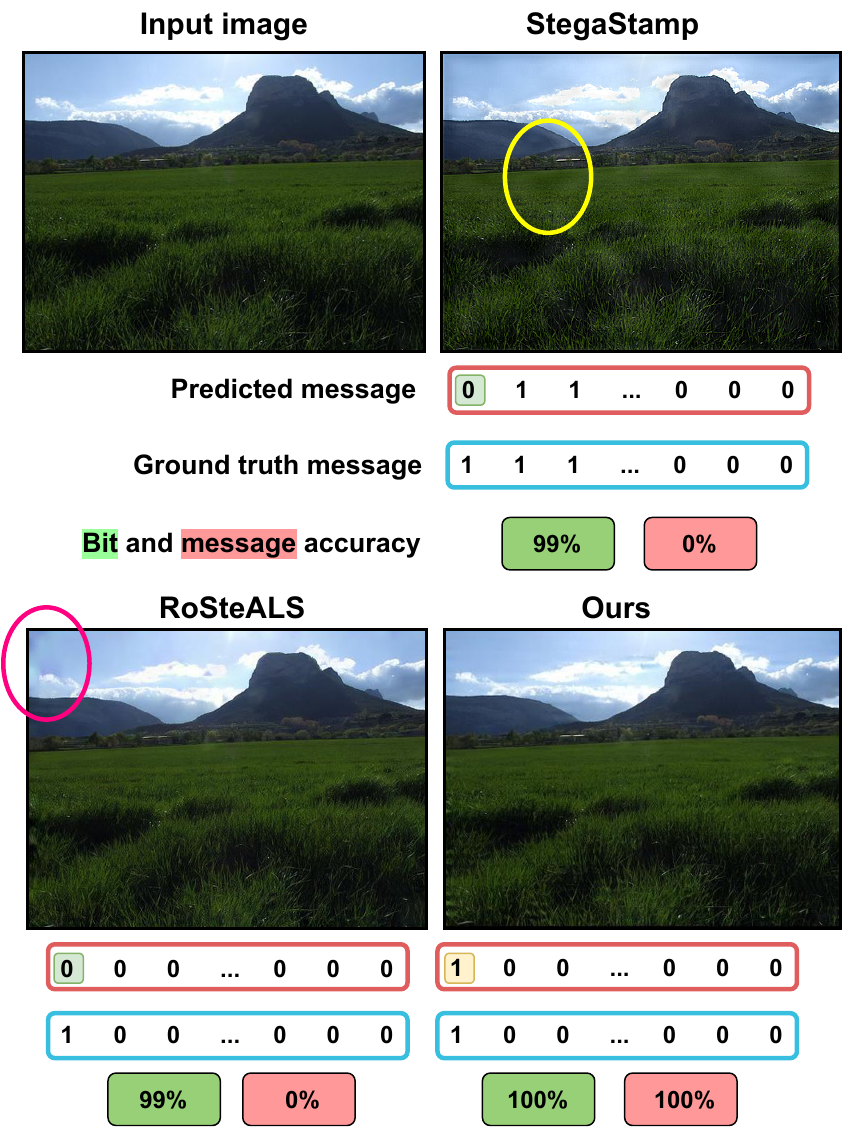}
    \caption{\textbf{Reconstructed images and messages of our method and prior work.} Although the image quality of all methods is similar, the message recovered in our approach has a much better \textbf{message accuracy} (our proposed metric) than those of prior work.}
    \label{fig:inference}
    \vspace{-10pt}
\end{figure}

In recent years, the emergence of advanced text-to-image models \cite{SD_T2I, Midjourney_T2I, Chang2023MuseTG_T2I, Nichol2021GLIDETP_T2I, Balaji2022eDiffITD_T2I, Ramesh2022HierarchicalTI_T2I, Saharia2022PhotorealisticTD_T2I} has underscored the pressing need to integrate these models with data-hiding techniques. These powerful image-generation tools provoke concerns regarding authorship protection and distinguishing between real and synthesized images. Watermarking stands out as a potential solution to address these concerns by enabling the embedding of hidden markers in synthesized images, facilitating author identification and highlighting their artificial origin. Additionally, steganography, beyond its role in secure information exchange, offers customizable watermarking to mitigate these issues. Consequently, within a remarkably short span, several papers have been published focusing on watermarking \cite{kim2023wouaf_t2iW, ma2023generative_t2iW, fernandez2023stable} and steganography \cite{bui2023rosteals} specifically tailored for text-to-image generation.

Despite the growing use of steganographic methods, previous evaluation frameworks \cite{zhu2018hidden, zhang1909rivagan, tancik2020stegastamp, bui2023rosteals} often neglect the vital need to completely preserve hidden secrets. These studies typically rely on the bit accuracy metric, which measures individual bit decoding but fails to assess the practical utility of image watermarking and steganography. For instance, a watermarking algorithm achieving 99\% bit accuracy may still fail to recover the entire message accurately, rendering it ineffective for confirming intellectual property ownership. Similarly, in steganography, even a high bit accuracy might overlook crucial errors in characters, impacting essential information like URLs, numbers, or encrypted text. Introduced alternative metrics, such as word accuracy as proposed by \citet{bui2023rosteals}, fail to capture the true usability of the system. This oversight calls for a re-examination of evaluation methodologies and a redefinition of criteria to better align with real-world requirements.

To bridge this gap, we propose a novel metric, \textbf{message accuracy}, as a departure from conventional evaluation methods like bit accuracy. This new metric only considers a match when the extracted hidden message perfectly aligns with the originally embedded content, emphasizing the crucial need for preserving entire hidden information in steganographic applications. Our findings using this metric reveal that state-of-the-art steganographic models, despite near-perfect bit accuracy, often demonstrate critically low practical value. For instance, the RoSteALS system, designed for 200-bit length messages, achieves a 94\% bit accuracy but rarely recovers an entirely correct message.

Moreover, conventional losses like MSE or BCE treat all predicted bits equally, which can saturate and provide uninformative guidance for network optimization when most bits are predicted correctly. To overcome this limitation, we introduce a novel loss function, the \textbf{``Log-Sum-Exponential'' (LSE) loss}. Unlike MSE or BCE, LSE prioritizes the gradient based on the most wrongly predicted bits, ensuring informative guidance for network optimization even with a few incorrect predictions. This approach notably enhances results in message accuracy. 

In addition, we introduce a novel technique leveraging the pre-trained Autoencoder of Stable Diffusion (SD) to embed confidential messages directly into synthesized images. Our approach involves a message encoder that aligns with the image content, ensuring more compatible message encoding and enhanced concealment capabilities without additional post-steganography steps.

We evaluate our methods against established approaches, RoSteALS \cite{bui2023rosteals} and StegaStamp \cite{tancik2020stegastamp}, for real image steganography across diverse datasets: MirFlickr \cite{huiskes2008mir}, CLIC \cite{mari2022clic}, and Metfaces \cite{karras2020metfaces}. Additionally, we conduct experiments in a generative setting, embedding hidden messages within newly generated images.

In summary, the contributions of our work are as follows:
\begin{enumerate}[leftmargin=7.5mm]
    \item We introduce \textbf{message accuracy}, a precise matching metric for extracted hidden messages.
    \item We propose a novel \textbf{LSE loss} enhancing recovery of complete hidden messages.
    \item We devise \textbf{latent-aware message encoding} using pretrained SD for steganographic image generation.
\end{enumerate}

In the following, Sec.~\ref{sec:related_work} reviews prior work; Sec.~\ref{sec:approach} specifies our approaches; and Sec.~\ref{sec:experiments} presents our implementation details and experimental results. Sec.~\ref{sec:conclusion} concludes with some remarks and discussions.

\label{sec:intro}



%% file: sec/2_related.tex
\section{Related Work}
\label{sec:related_work}
A large number of techniques have been developed for image-based steganography. They can be categorized as classical, deep-learning, and generative-based methods. 

\myheading{Classical methods.}
The Least-Significant Bit (LSB) embedding method, an early hand-crafted technique \cite{Wolfgang1996AWF}, concealed data within the lowest bits of image pixels to create steganographic images visually akin to originals. Subsequent advancements, including spatial \cite{ghazanfari2011lsb++, taha2022high} and frequency domain techniques \cite{holub2012designing, holub2014universal, li2007steganographic, navas2008dwt, pevny2010using, provos2001defending}, aimed to increase capacity while preserving visual quality. Despite advantages, these methods rely on delicate features, making them vulnerable to even minor alterations that could lead to significant hidden information loss.


\myheading{Deep-Learning-based methods.}
Deep learning has revolutionized steganography, enhancing accuracy, imperceptibility, and resilience. HiDDeN \cite{zhu2018hidden} pioneered an end-to-end framework employing an encoder-decoder architecture with a transformation layer and adversarial discriminator. Subsequent advancements in architectures \cite{chang2021neural, duan2019reversible, huang2022image, meng2018fusion, tancik2020stegastamp, zhang1909rivagan} continue to refine steganographic image quality and robustness, often emphasizing joint encoding of secrets and covers. Our approach diverges, leveraging a latent space generated by a pretrained autoencoder within Stable Diffusion, offering increased resilience to image transformations compared to direct RGB space embedding.

\myheading{Generative-based methods.} With advancements in text-to-image generative models like Stable Diffusion (SD) \cite{rombach2022highLatent}, new methods embedding watermarking/steganography have emerged. Stable Signature \cite{fernandez2023stable} fine-tunes SD's image decoder for steganographic image generation, but it necessitates re-finetuning for each new message. RoSteALS \cite{bui2023rosteals} explores VQGAN's latent representations for concealing information but lacks image-content-aware message encoding. In contrast, our approach incorporates image latent information into the message encoder, ensuring compatibility and preserving concealed information.

\myheading{Evaluation metrics for steganography.}
Two metric sets gauge steganography effectiveness: one assessing stealthiness through visual similarity (e.g., PSNR, SSIM, LPIPS), the other measuring reconstructed secret accuracy, commonly using bit accuracy like HiDDeN \cite{zhu2018hidden}. RoSteALS \cite{bui2023rosteals} introduces ``word accuracy'', evaluating decoded messages with at least 80\% bit accuracy. However, none fully assesses decoded secrets, missing watermarking and steganography practical goals. To address this, we propose ``message accuracy'', evaluating entire decoded messages. Additionally, we present a novel, adaptable loss function to enhance message accuracy and bolster practical system performance.

%% file: sec/3_approach.tex
\section{Proposed Approach}
\label{sec:approach}

\myheading{Problem Statement:}
In steganography, user A wants to conceal a hidden message in the form of a binary string $m \in \{0, 1\}^{d}$ with length $d$ within a cover image $I \in \mathbb{R}^{H \times W \times 3}$ ($H$ and $W$ are the image height and width) using a message encoder $\E$, ensuring that the altered image $I'$ is visually similar to original image $I$. Subsequently, user B equipped with the message decoder $\D$ provided by user A can extract the hidden message $m'$. We expect to transfer the message without any information loss, i.e., $m' = m$.

To address this problem, we make three key contributions. Firstly, we propose a strict metric known as \textbf{message accuracy}, emphasizing bit-wise identical between the extracted hidden message and the originally embedded message. Secondly, to address the strict requirements of the message accuracy metric, we introduce a novel \textbf{LSE loss} aimed at improving the recovery of the entire hidden message. Thirdly, we present a technique for \textbf{latent-aware message encoding} based on Stable Diffusion for generating steganographic images.

\begin{figure*}
    \centering
    \includegraphics[width=1.\linewidth]{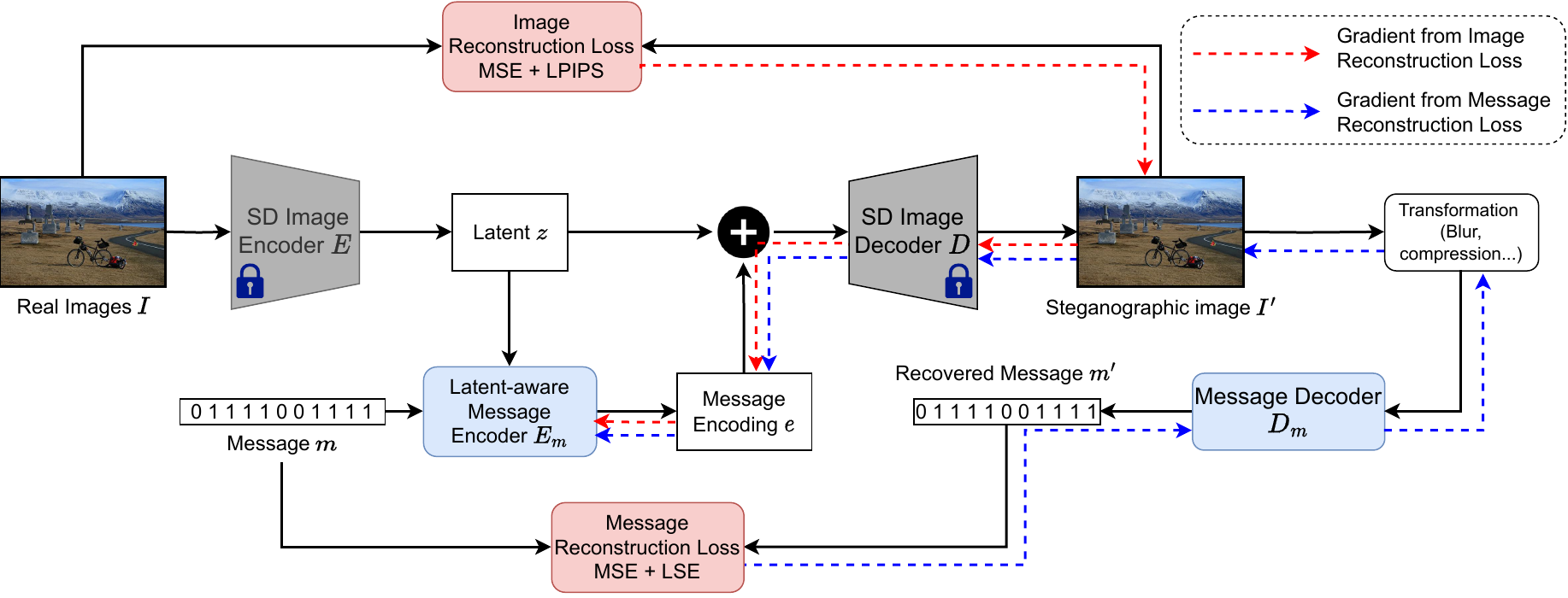}
    \caption{\textbf{Training of \Approach.} Given a real image $I$, we utilize the Image Encoder $E$ of SD to extract its latent $z$. Subsequently, the Latent-aware Message Encoder $E_m$ takes as input the message $m$ and latent $z$ to produce the message encoding $e$. Next, the SD Image Decoder $D$ receives the modified latent $z'=z+e$ to generate steganographic image $I'$. Optionally, $I'$ can be further transformed with some operations such as blur and compress to enhance the robustness. Finally, the Message Decoder $D_m$ recovers the hidden message $m'$ in $I'$. To train the network, we use two sets of loss functions: image reconstruction and message reconstruction.}
    \label{fig:main-scheme}
\end{figure*}

\begin{figure}
    \centering
    \includegraphics[width=1\linewidth]{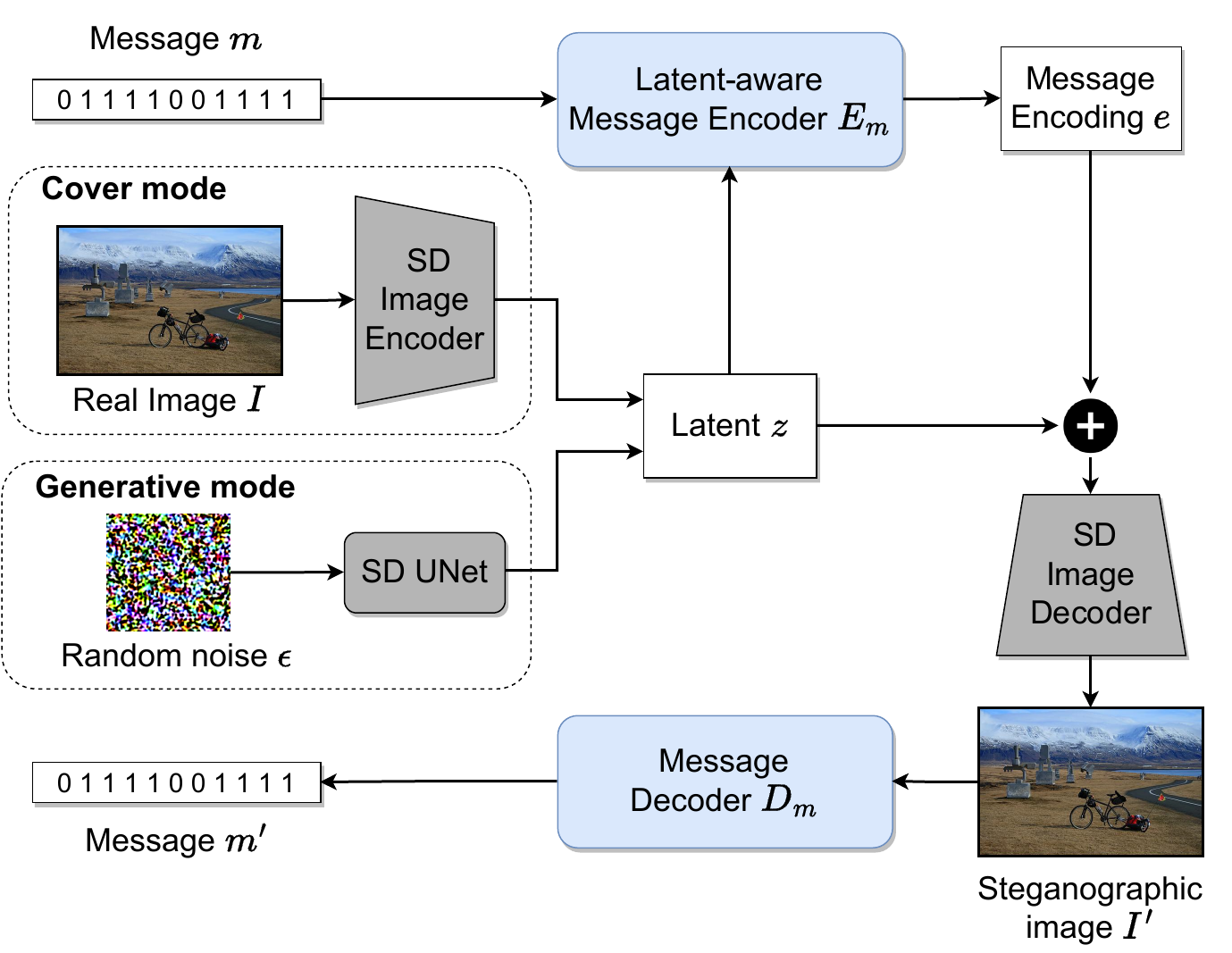}
    \caption{\textbf{Testing of \Approach.} There are two modes: cover mode and generative mode. In the cover mode, the real image $I$ is encoded into latent $z$ with a pretrained SD Image Encoder while in the generative mode, the pretrained UNet of SD transforms a noise $\epsilon$ to latent $z$. Subsequent steps are similar to those in the training of \Approach. See more in caption of Fig.~\ref{fig:main-scheme}.}
    \label{fig:inference}
\end{figure}

\subsection{Message Accuracy Metric}
Previous watermarking/steganography works \cite{zhu2018hidden, zhang1909rivagan, tancik2020stegastamp, fernandez2023stable, bui2023rosteals} utilized \textit{bit accuracy} computed as:
\begin{align}
    \text{Bit Accuracy }(m, m') = \frac{1}{d}\sum_{i=1}^d \mathbbm{1}(m_i = m'_i), 
\end{align}
where $\mathbbm{1}(\cdot)$ is indicator function, and $m_i, m'_i$ are the $i$-th bits of $m, m'$.
The common practice of focusing solely on the bit accuracy metric may inadvertently overlook the critical need for holistic message accuracy. High bit accuracy, such as achieving 99\%, does not inherently guarantee the exact extraction of the entire message, as can be observed in our comparison results (Tab.~\ref{tab:comparison}). This oversight raises concerns about the reliability of steganographic techniques, as a seemingly high bit accuracy rate may still result in critical portions of the concealed message being inaccurately retrieved. Such a discrepancy could be especially problematic when the extracted information is used for legal or security. 
Thus, introducing and emphasizing the \textbf{message accuracy} metric becomes crucial and necessary for developing a reliable steganography system. A shift toward evaluating methods focusing on the comprehensive accuracy of the entire hidden message ensures a more robust and dependable assessment, aligning steganography research with real-world applicability and the stringent demands of secure information retrieval.
Formally, message accuracy is calculated as:
\begin{align}
    \text{Message Accuracy }(m, m') = \bigwedge_{i=1}^d \mathbbm{1}(m_i = m'_i), 
\end{align}
where $\wedge$ is the AND operator. Equivalently, a message is correct if and only if all of its bits are correct.

\subsection{Loss-Sum-Exponential (LSE) Loss}
To tackle the stringent requirement of the message accuracy metric, we introduce a new loss using the Log-Sum-Exponential function, named the \textbf{LSE loss}. For an input $x \in \mathbb{R}^{d}$, the LSE between predicted message $m$ and ground-truth message $m^*$ is computed as:
\begin{equation}
    \mathcal{L}_{\text{LSE}}(m, m^*) = \log{\left( \sum_{i=1}^{d}\exp \|m_i - m^*_i\|_2^2 \right)}.
\end{equation}

Let us explain why the LSE loss helps improve message accuracy while the BCE and MSE loss functions do not. The loss of BCE or MSE becomes smaller when \textit{most} of the bits are predicted correctly, and thus, its gradient becomes uninformative for the network's optimization. In other words, a few bits incorrectly predicted do not affect the loss much as the loss is averaged across all bits. This is against the requirement of the message accuracy metric that all bits of the retrieved message must be correct. 
However, the Log-Sum-Exponential function in the LSE loss can be considered as a ``soft''-max function where the maximum bit discrepancy at a position dominates the entire loss. Hence, it maintains informative gradients for updating the model's parameters even if only a few bits are predicted wrongly. 

Nevertheless, it should be cautious when using the LSE loss at the beginning of the training process. That is, when most bits are not correct, the LSE loss can ``explode''. Thus, it is advised to use LSE in conjunction with BCE or MSE when the model reaches a certain level of bit accuracy. It is noteworthy that the new LSE loss is applicable to various kinds of steganography approaches, not only ours.

\subsection{\Approach}

\myheading{Architecture:}
To begin, let us depict the training process of \Approach~in Fig.~\ref{fig:main-scheme}, which is based on the pretrained Stable Diffusion \cite{rombach2022highLatent}.
First, we obtain the latent code $z$ from the cover image $I$ using the Image Encoder $\E$ of SD. Next, the Latent-aware message encoder $\E_m$ computes the message encoding $e$ from the given message $m$ and latent $z$, which is subsequently added to the original latent $z$ to produce the steganographic latent $z' = e + z$. Then, the Image Decoder $\D$ of SD takes as input $z'$ to generate the steganographic image $I'$. Finally, the message decoder $\D_m$ extracts the hidden message $m'$ concealed inside $I'$. 
Note that the message encoder and decoder can be trained jointly. Optionally, steganographic image $I'$ can be corrupted by some differentiable transformation operations, e.g., blur and compression, to enhance the robustness.

\begin{table*}[t]
    \centering
    \begin{tabular}{llccccl}
    \toprule
     \bf Datasets & \bf Methods & \bf PSNR \(\uparrow \)& \bf SSIM \(\uparrow\)& \bf LPIPS \( \downarrow\) & \bf Bit Acc (\%) \(\uparrow\) & \bf Message Acc (\%) \(\uparrow\) \\ 
    \toprule 
    \multirow{ 6}{*}{\textbf{MirFlickr \cite{huiskes2008mir}}} & StegaStamp \cite{tancik2020stegastamp}   & 29.74     &0.91      &  0.07                          &   99           &      87         \\
    & StegaStamp \cite{tancik2020stegastamp} + LSE   & 29.97   & 0.92      &0.07                         &  99          &       96 \textbf{(+9)}            \\
     & RoSteALS \cite{bui2023rosteals}      & 33.16      & 0.93      &0.04 & 99   & 67                  \\
    & RoSteALS \cite{bui2023rosteals} + LSE     &  33.22      & 0.94   &  0.04   &    99      & 73 \textbf{(+6)}                \\
    & Ours &    33.31  &0.87  &0.07  & 99  & 77  \\
    & Ours + LSE    &33.11  &0.87  &0.07  & 99  & 94 \textbf{(+17)} \\
    \toprule 
    \multirow{ 6}{*}{\textbf{CLIC \cite{mari2022clic}}} & StegaStamp \cite{tancik2020stegastamp}    & 30.37     &0.94      &  0.07                          &   99           &      92            \\
    & StegaStamp \cite{tancik2020stegastamp} + LSE   & 30.66     &0.94      &  0.07                         &  99          &       97 \textbf{(+5)}   \\
     & RoSteALS \cite{bui2023rosteals}      &33.41       & 0.97      & 0.04 & 99   & 71                 \\
    & RoSteALS \cite{bui2023rosteals} + LSE     & 33.47     & 0.97      & 0.04 & 99   & 78 \textbf{(+7)}                  \\
    & Ours     & 33.60  &0.93  &0.08  &  99  & 75  \\
    & Ours  + LSE   &33.27  &0.93  &0.08  & 99  & 95 \textbf{(+20)}  \\
    \toprule 
    \multirow{ 6}{*}{\textbf{Metfaces \cite{karras2020metfaces}}} & StegaStamp \cite{tancik2020stegastamp}    & 32.04     &0.95      &  0.12                          &   99           &      96            \\
    & StegaStamp \cite{tancik2020stegastamp} + LSE   & 32.48     &0.95      &  0.12                          &   99          &       99 \textbf{(+3)}           \\
     & RoSteALS \cite{bui2023rosteals}      & 35.84      &0.97      &0.06 & 99   & 75                 \\
    & RoSteALS \cite{bui2023rosteals} + LSE     & 35.89      &0.98      &0.06   & 99   & 80 \textbf{(+5) }                 \\
    
    & Ours     &35.29  &0.92  &0.11  & 99  & 91  \\
    & Ours + LSE  &35.23  &0.92  &0.11  & 99  & 99 \textbf{(+8)}\\ 
    \bottomrule
    \end{tabular} 
    \caption{Results of our \Approach~and prior work with and without the LSE loss on various datasets using \textbf{100-bit message}.
    }
    \vspace{-10pt}
    \label{tab:comparison}
\end{table*}

The main reason is that we want our approach to work with both \textbf{cover mode} (hide a message inside a real image) and \textbf{generative mode} (embed a message in the image generation process given a text prompt) in testing, we propose to use the image encoder and decoder of Stable Diffusion (SD) \cite{rombach2022highLatent} as illustrated in Fig.~\ref{fig:inference}. In particular, in the cover mode, the cover image can be encoded using the image encoder to a latent $z$ while in the generative mode, the UNet of SD can be used to iteratively transform a noise $\epsilon \sim \mathcal{N}(0, I)$ to the final latent $z$. It is worth noting that the latent code $z$ has the size of $\mathbb{R}^{\frac{H}{8}\times \frac{W}{8} \times 4}$ capturing the sufficient content of the image.
This architecture is in contrast to prior work, RoSteALS \cite{bui2023rosteals} where the message encoding is created independently from the image content. In other words, the message encoder $\E_m$ is not aware of the image content so the message encoding might not be compatible with the image, resulting in information loss in the steganographic image. Therefore, we propose to take the latent code $z$ which captures the essential content of the image along with the message $m$ as inputs to the latent-aware message encoder $\E_m$. The effectiveness of the latent-aware message encoder will be demonstrated in the Experiment section. 

\myheading{Training loss functions:}
As mentioned above, we follow two sets of training loss functions including image reconstruction and message reconstruction. For image reconstruction loss, we follow RoSteALS \cite{bui2023rosteals} to use LPIPS loss \cite{zhang2018lpips} and MSE. For the message reconstruction loss, we additionally use the LSE loss at a particular iteration $t$ along with the MSE loss. The final loss we use to train our \Approach is as follows:
\begin{align}
    \mathcal{L} = \alpha_1 \mathcal{L}_{\text{LPIPS}}^{\text{image}} + \alpha_2 \mathcal{L}_{\text{MSE}}^{\text{image}} + 
    \alpha_3 \mathcal{L}_{\text{LSE}}^{\text{message}} + 
    \alpha_4
    \mathcal{L}_{\text{MSE}}^{\text{message}}.
\end{align}
One advantage of our method is that we only train in the cover mode to speed up the training process but can test in both cover and generative modes for flexible usage.

%% file: sec/4_experiments.tex
\section{Experiments}
\label{sec:experiments}

\subsection{Experimental Setup}
\myheading{Datasets:}
We conduct our experiments on three datasets: MirFlickr \cite{huiskes2008mir}, CLIC \cite{mari2022clic}, and MetFaces \cite{karras2020metfaces}.  In our experiments, we \textit{train} on 100K real images and \textit{validate} in two modes: cover and generative modes. For the cover mode, we tested on another set of 1,000 images of MirFlickr, 530 test images of CLIC, and 1336 test images of MetFaces. For the generative mode, we utilize the image captions of the 1,000 images of the Flickr 8K dataset \cite{hodosh2013framing}. 
%

\myheading{Evaluation metrics:} We use two sets of evaluation metrics: image quality including PSNR \cite{korhonen2012peak}, SSIM \cite{wang2004ssim}, and LPIPS \cite{zhang2018lpips}, and message preservation encompassing Bit accuracy and Message accuracy.

\myheading{Implementation details:}
During training, the input image is resized to 512 $\times$ 512 and then fed to the Image Encoder of SD \cite{rombach2022highLatent}. We experiment with message lengths of 100 bits. We use the AdamW optimizer with a learning rate of $8e^{-5}$. By default, we set $\alpha_1=1.0, \alpha_2=1.5, \alpha_3=0.1, \alpha_4=16.0$. We also follow RoSteALS's \cite{bui2023rosteals} strategy to stabilize the training. In particular, we start with a fixed image batch and then unlock the full training data after bit accuracy reaches a threshold $\tau_1 = 90\%$. After training on the full dataset and waiting for the bit accuracy to reach the threshold $\tau_2 = 95\%$, we apply image transformation to make the method more robust to the transformed inputs in testing. Empirically, we also activate the LSE loss when the bit accuracy reaches the threshold $\tau_2$.
For the architecture of message encoder $\E_m$, first, the 1D message $m$ is converted to a 2D message using 1 fully connected layer and subsequently concatenated with the latent $z$. The output is then taken as input to a UNet architecture with 4 down and 4 up layers to produce the message encoding $e$.


\begin{figure*}
    \centering
    \includegraphics[width=1\linewidth]{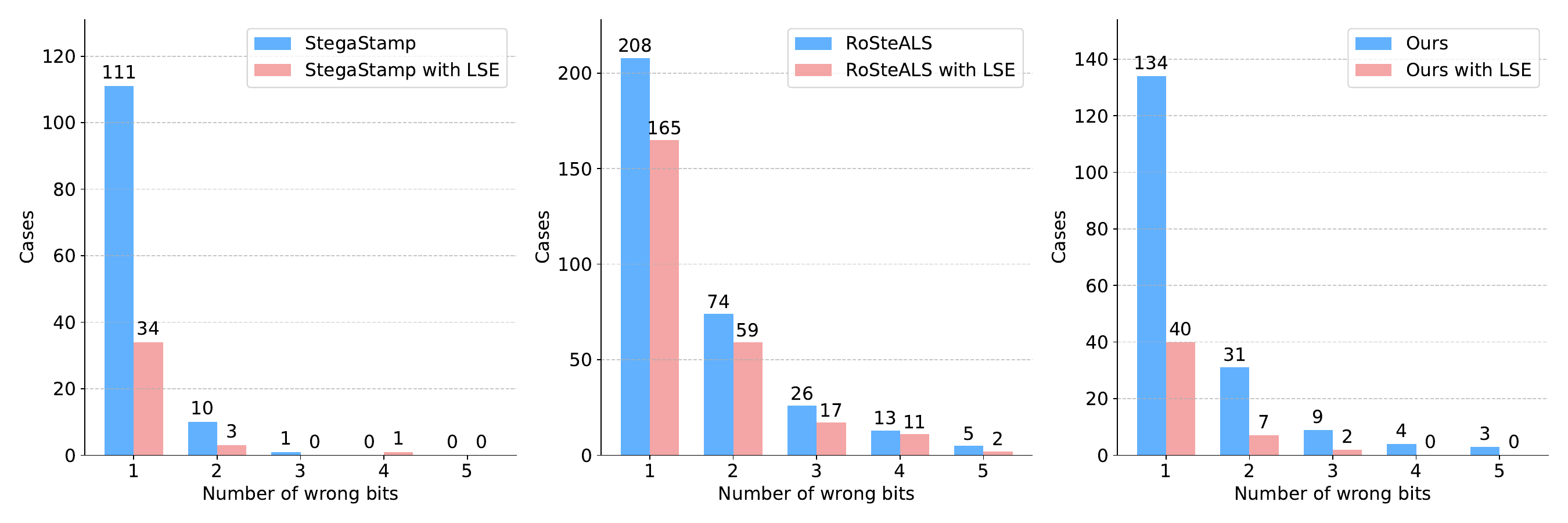}
    \caption{Histograms of wrong bits with and without using the LSE loss in StegaStamp \cite{tancik2020stegastamp}, RoSteALS \cite{bui2023rosteals}, and ours.} 
    \label{fig:wrong-bits}
\end{figure*}

\subsection{Comparison with Prior Methods}
We compare our approach with the following baselines:
\begin{itemize}
    \item RoSteALS \cite{bui2023rosteals}: a method uses the latent representation produced by the pretrained autoencoder of VQGAN \cite{esser2021taming} to conceal the message. We use the official code\footnote{\url{https://github.com/TuBui/RoSteALS}} to reproduce the results.
    \item StegaStamp \cite{tancik2020stegastamp}: a deep learning-based method. We also use the official code\footnote{\url{https://github.com/tancik/StegaStamp}} to reproduce the results.
\end{itemize}

\myheading{Cover mode.}
We present the comparative results between our approach and prior work on the three datasets MirFlickr \cite{huiskes2008mir}, CLIC \cite{mari2022clic}, and Metfaces \cite{karras2020metfaces} in Tab.~\ref{tab:comparison}. We can see that StegaStamp \cite{tancik2020stegastamp} usually has better message accuracy but lower image quality. RoSteALS \cite{bui2023rosteals} is a latent-based approach with higher image quality but lower message accuracy. \Approach~is also a latent-based approach that enjoys a better trade-off, i.e., having high image quality as RoSteALS but comparable message accuracy to StegStamp thanks to the proposed latent-aware message encoder. 

Furthermore, the performance gains of using the LSE loss indicate that our LSE loss exhibits robustness across various deep-learning-based steganography methods. Applying our proposed loss consistently leads to a notable increase in message accuracy without discernible degradation in image quality. To explain why using our LSE loss helps improve the message accuracy a lot, we count the number of cases with wrong bits in each message and present them in the histogram in Fig.~\ref{fig:wrong-bits}. The number of wrong bits reduced significantly after applying the LSE loss, resulting in better message accuracy.

\begin{table}[!t]
    \small
    \setlength{\tabcolsep}{7pt}
    \centering
    \begin{tabular}{lccc}
    \toprule 
         \textbf{Approach A} &  \bf PSNR $\uparrow$ & \bf Bit Acc $\uparrow$ & \bf Message Acc  $\uparrow$ \\
            \midrule 
          RoSteALS \cite{bui2023rosteals}   & 38.90   & 98\% &  48\%   \\ 
         Ours    & 35.18    & 99\% & 89\%  \\ 
         \toprule
    \end{tabular}
    \caption{Results of applying steganography in the process of generating new images.}
    \label{tab:generative-mode}
\end{table}

\begin{table}[t]
    \small
    \setlength{\tabcolsep}{2pt}
    \centering
    \begin{tabular}{lcccc}
    \toprule
        \textbf{Approach B} &  \bf PSNR $\uparrow$   & \bf SSIM $\uparrow$  & \bf LPIPS $\downarrow$ &   \textbf{Message Acc} $\uparrow$  \\
         \midrule
         RoSteALS \cite{bui2023rosteals}                &33.79  &0.96  &0.04  & 73\%  \\
         StegaStamp \cite{tancik2020stegastamp} &32.84  &0.92  &0.08  & 90\%  \\
         Ours        &34.37  &0.90  &0.04  & 93\% \\
    \toprule 
    \end{tabular}
    \caption{Results of applying steganography on generated images.}
    \label{tab:stega-on-generated}
    \vspace{-10pt}
\end{table}

\myheading{Generative mode.}
In Table~\ref{tab:generative-mode}, we present a comparison between our network's performance and RoSteALS in concealing a message during the image generation process. As RoSteALS \cite{bui2023rosteals} does not provide a pretrained message encoder-decoder in this mode, we had to re-implement it using the official code's message encoder-decoder. We also retained the latent $z$ to generate the reference image for evaluating the image quality. The quantitative results demonstrate that our network is more effective than RoSteALS in recovering the hidden message, albeit at a slight cost to image quality.

Another way to apply steganography in generative mode is to apply steganography approaches to the generated images as cover images. The results in Tab.~\ref{tab:stega-on-generated} suggest that our method achieves a better message accuracy, surpassing RoSteALS \cite{bui2023rosteals} with a gap of 20\%. Compared to StegaStamp \cite{tancik2020stegastamp}, we have slightly better results in message accuracy while delivering better image quality.
Moreover, the question that may arise is whether to apply steganography in the process of generating images (approach A) or to generate images first and then apply steganography (approach B). Comparing the results in Tab.~\ref{tab:generative-mode} and Tab.~\ref{tab:stega-on-generated}, we can conclude that when the image quality is preferable, use approach A, otherwise, use approach B.

\begin{figure*}
    \centering
    \includegraphics[width=1\linewidth]{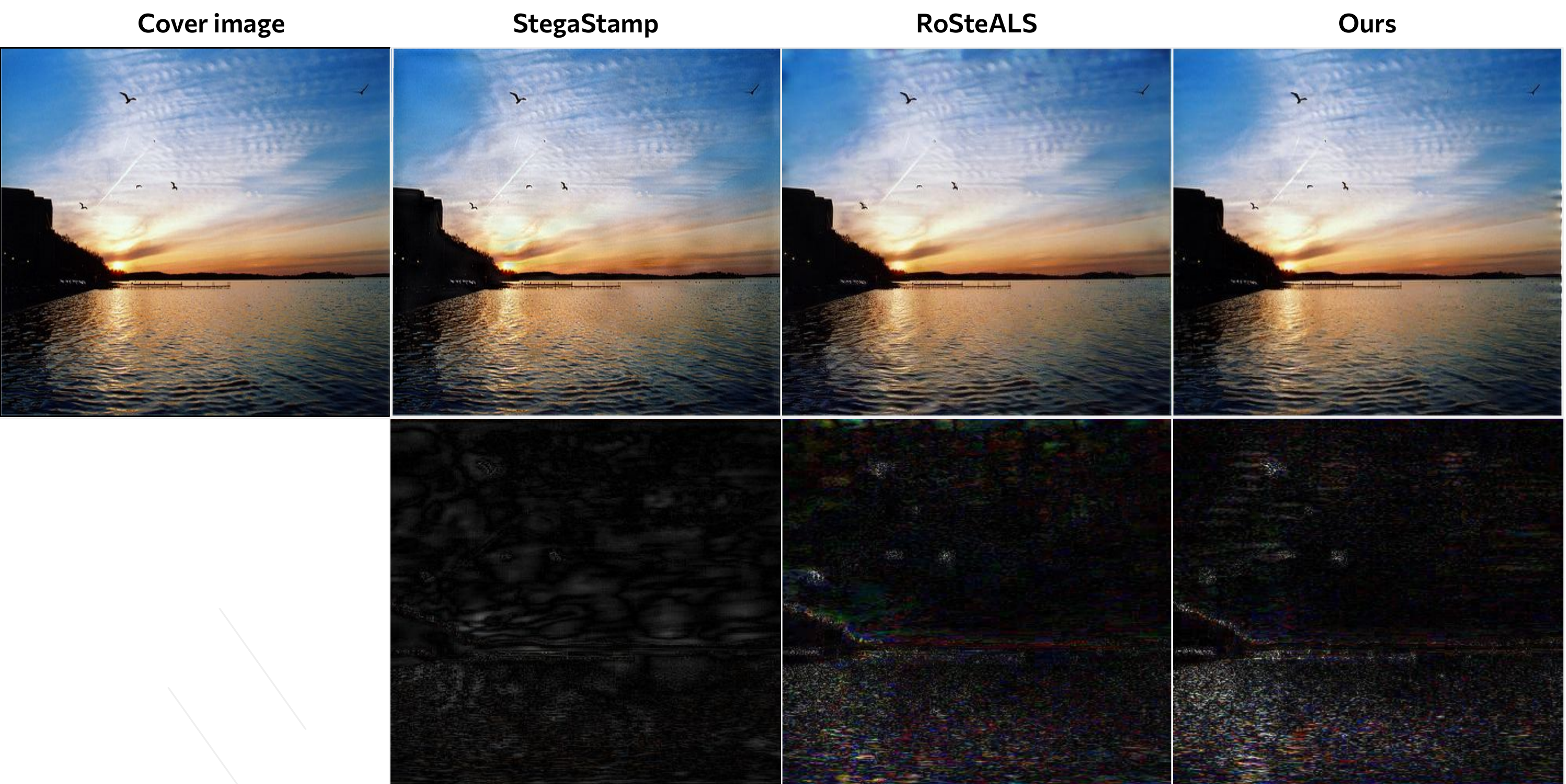}
    \caption{\textbf{Qualitative results in the cover mode.}
    Different methods create their artifact on the cover image. The second row shows the residual between the cover image and the steganographic image (the residual is magnified $2\times$ for visualization purposes only).}
    \label{fig:cover-mode}
\end{figure*}

\begin{figure*}
    \centering
    \includegraphics[width=1\linewidth]{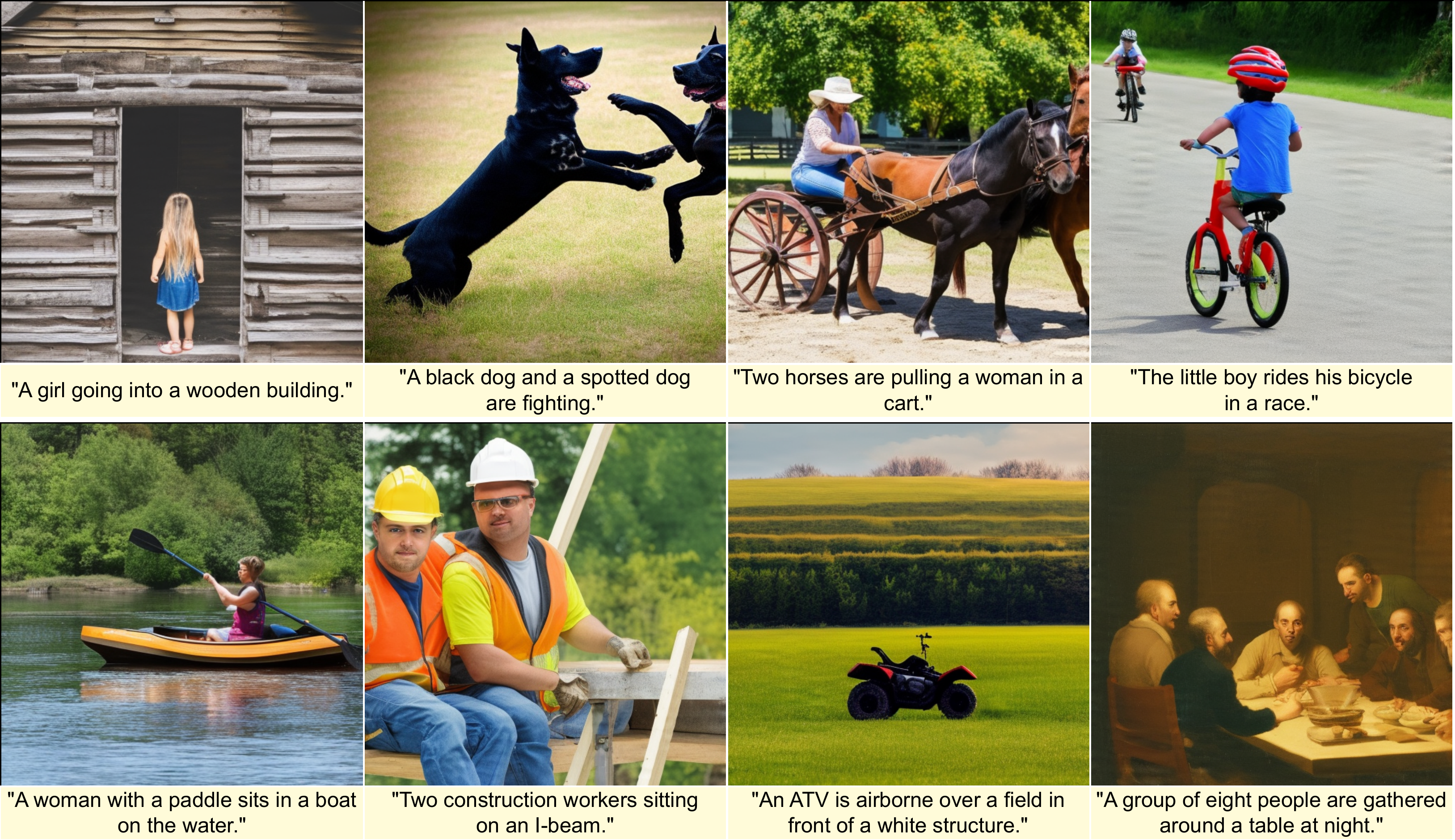}
    \caption{\textbf{Qualitative results in the generative mode}. Our method generates high-fidelity steganographic images (first three columns) but occasionally reveals hidden patterns in the resulting images (last column).}
    \label{fig:generative-mode}
\end{figure*}

\begin{table*}[t]
    \small
    \centering
    \begin{tabular}{lcccccc}
    \toprule
    & \multicolumn{2}{c}{\textbf{StegaStamp \cite{tancik2020stegastamp}}} & \multicolumn{2}{c}{\textbf{RoSteALS \cite{bui2023rosteals}}} & \multicolumn{2}{c}{\textbf{Our + LSE}} \\
 & \bf Bit Acc  (\%)    & \bf Message Acc (\%)     & \bf Bit Acc (\%)   &  \bf Message Acc (\%)   & \bf Bit Acc (\%) & \bf Message Acc (\%)\\
 \toprule
  No Transformation  & 99  & 87  & 99  & 67  & 99  & \textbf{94} \\ 
 Gaussian blur           & 99 & 89    &                     99&                 61 &                   99& \textbf{94}              \\
 Gaussian noise          &  92  &  1 &                     93&                 2&                   99& \textbf{46}             \\
 RGB2BGR                 & 99 & 86 &                     99&                 53&                   99&   \textbf{92}              \\
JPEG Compression                 &   99  &  85 &                     99&                 60&                   99&   \textbf{93}             \\
\bottomrule
\end{tabular}
    \caption{Robustness results on the MirFlickr \cite{huiskes2008mir} dataset under various color intensity transformations.}
    \vspace{-10pt}
    \label{tab:robust-evaluation}
\end{table*}

\myheading{Qualitative comparison.}
We present qualitative results for the cover mode in Fig.~\ref{fig:cover-mode} and generative mode in Fig.~\ref{fig:generative-mode}. For the cover mode, each method introduces a distinct visual artifact to the cover image. For the generative mode, the first two columns exhibit several high-fidelity samples, while the last column showcases failure cases, where observable visual patterns emerge.


\subsection{Robustness Evaluation}
We aim to assess the real-world resilience of our steganographic method, comparing its performance against RoSteALS \cite{bui2023rosteals} and StegaStamp \cite{tancik2020stegastamp}. The experiment introduces various color intensity transformations, simulating potential modifications that steganographic images may undergo in practical scenarios. 
To this end, we use the provided checkpoint of RoSteALS \cite{bui2023rosteals}, and StegaStamp \cite{tancik2020stegastamp} to compare with our proposed method. We evaluate the robustness of these steganography methods with 1000 images taken from MirFlickr \cite{huiskes2008mir}. Additionally, we discuss the hyper-parameters of each transformation in the supplementary document. From Tab.~\ref{tab:robust-evaluation}, we found that our approach is robust with several modifications such as Gaussian blur, transform color, or jpeg compression with a modest drop in message accuracy. However, the message accuracy significantly degrades if the steganographic images are injected with Gaussian noise. Nonetheless, ours still outperforms others with significant margins.

\subsection{Ablation Study}
Noticeably, at the beginning of the training of our approach, we only use MSE loss as our message reconstruction loss, later, after the bit accuracy reaches some threshold $\tau_1$, we activate the LSE loss. Thus, we study the impact of LSE coefficient $\alpha_3$ and message MSE coefficient $\alpha_4$ using the test set of MirFlickr \cite{huiskes2008mir}. Additionally, we provide more ablation studies for network design in the supplementary.

\myheading{Study on LSE coefficient $\alpha_3$} is shown in Tab.~\ref{tab:ablate_lse_w}. As observed, when the coefficient is too small, it results in a low message accuracy. Our selected value of $\alpha_3 = 0.1$ yields the best results.

\myheading{Study on the message MSE coefficient $\alpha_4$} is shown in the Tab.~\ref{tab:ablate_sw}. Our observations reveal that employing a higher value contributes to incrementally improved message accuracy with an acceptable trade-off in PSNR. Notably, it is essential to acknowledge that a higher value also accelerates the increase in bit accuracy, reaching the first threshold $\tau_1 = 90\%$ faster. Therefore, we choose $\alpha_4 = 16$. 

\begin{table}[t]
    \small
    \centering
    \begin{tabular}{lcccc}
    \toprule
         $\alpha_3$ &  \textbf{0.025} & \textbf{0.05}   & \textbf{0.1}    & \textbf{0.2} \\
         \midrule
         Bit Acc (\%)    & 99      & 99      & 99      & 99 \\  
         Message Acc (\%)   & 73      & 71      & 94      & 94 \\  
    \toprule
    \end{tabular}
    \caption{Study on LSE loss's coefficient \(\alpha_3\)}
    \label{tab:ablate_lse_w}
\end{table}

\begin{table}[t]
    \small
    \centering
    \begin{tabular}{lcccc}
    \toprule
         $\alpha_4$ &  \textbf{10} & \textbf{12}    &\textbf{14 }   &\textbf{16} \\
         \midrule
         PSNR           &33.16          &33.18      &33.30          &32.73 \\ 
         Bit Acc (\%)    & 99      & 99      & 99      & 99 \\  
         Message Acc (\%)   & 94      & 97      & 96      & 99 \\   
         \# Iterations to pass $\tau_1$    & 41K      &34K     & 31K      &29K \\ 
    \toprule
    \end{tabular}
    \caption{Study on message coefficient $\alpha_4$.}
    \vspace{-10pt}
    \label{tab:ablate_sw}
\end{table}

%% file: sec/5_conclusion.tex
\section{Discussion}
\label{sec:conclusion}

\myheading{Limitations:} 
Even with its strength against color intensity shifts, our approach faces a challenge: geometric transformations like rotation or perspective warp. Unlike color alterations, these transformations can disrupt the spatial relationships in the image, jeopardizing the integrity of the encoded message's hidden pattern within steganographic images.
Future research could focus on developing adaptive techniques that maintain the encoded message's integrity despite geometric distortions.

\myheading{Conclusion:}
In conclusion, we have made three key contributions. First, we have introduced a novel metric of message accuracy, emphasizing precise alignment between the extracted hidden message and the originally embedded message. Second, we have devised a novel LSE loss to significantly enhance the entire message recovery, meeting the strict requirements of the message accuracy metric. Next, we have proposed our approach \Approach~with the latent-content-aware message encoding technique leveraging a pretrained SD, contributing to a better trade-off between image quality and message recovery. Most importantly, our approach can work in both cover and generative modes, where the latter is very crucial for protecting and verifying generated photo-realistic images from very powerful text-to-image generative models.